\documentclass{article} 
\usepackage{nips15submit_e,times}
\usepackage{hyperref}
\usepackage{url}
\pdfoutput=1

\usepackage{graphicx}
\usepackage{subfigure}
\usepackage[small]{caption}
\usepackage{cite}
\usepackage{booktabs}

\usepackage{amsmath}

\usepackage{tabularx}
\usepackage{microtype}

\usepackage{tikz,times}
\usetikzlibrary{positioning,decorations.pathmorphing}
\usepackage{geometry}
\usetikzlibrary{mindmap,backgrounds}
\usetikzlibrary{arrows,shapes}

\title{Efficient Empowerment}

\author{Maximilian Karl, Justin Bayer, Patrick van der Smagt\\
       Technische Universit\"at M\"unchen\\
       karlma@in.tum.de, bayer.justin@googlemail.de, smagt@brml.org\\ 
}

\nipsfinalcopy 

\begin{document}

\maketitle

\begin{abstract}

Empowerment quantifies the influence an agent has on its environment. This is formally achieved by the maximum of the expected KL-divergence between the distribution of the successor state conditioned on a specific action and a distribution where the actions are marginalised out. This is a natural candidate for an intrinsic reward signal in the context of reinforcement learning: the agent will place itself in a situation where its action have maximum stability and maximum influence on the future. The limiting factor so far has been the computational complexity of the method: the only way of calculation has so far been a brute force algorithm, reducing the applicability of the method to environments with a small set discrete states.
In this work, we propose to use an efficient approximation for marginalising out the actions in the case of continuous environments. This allows fast evaluation of empowerment, paving the way towards challenging environments such as real world robotics. The method is presented on a pendulum swing up problem.

\end{abstract}

\section{Introduction}

Empowerment \cite{klyubin_keep_2008, salge_empowerment_2013} is an information theoretic quantity measuring the amount of information induced by an agents actuators and the information perceived by its sensors.
It therefore measures the amount of control over the environment but also how well the current state can be perceived by the sensors.
\cite{jung_empowerment_2012, klyubin_keep_2008} showed that system states with high empowerment value have maximal future options.
In the case of an inverted pendulum this state with maximum future possibilities consists of balancing the pendulum in an upright position as shown in experiments using the empowerment formulation \cite{jung_empowerment_2012}.
This value can be used in reinforcement learning as the reward function and serves as an unsupervised type of control which moves the robot towards states with high stability and maximal influence.

Previous applications lack an efficient implementation and the ability to use continuous variables either for the state space or the action space.
They do not scale well with the dimension of the action space which limits empowerment to simple simulations.
The very first implementations assumed discrete distributions for both spaces \cite{klyubin_keep_2008} and later \cite{jung_empowerment_2012} used empowerment for continuous states but still needs a low dimensional discrete action space.
Real-world robotics tasks, such as in-hand manipulation, would require a high dimensional continuous action space.
We developed an efficient computation of empowerment able to cope with high dimensional continuous state and action spaces enabling the use of empowerment for real-world robotic tasks.

\section{Empowerment}

Empowerment $C(x)$ is defined as the Shannon channel capacity\cite{jung_empowerment_2012}:

\[
        C(x) := \operatornamewithlimits{max}_{p(a|x)} \int p(a|x) \int p(x^{\prime} | x, a) \ln \frac{p(x^{\prime} | x, a)}{p(x^{\prime} | x) } dx^{\prime} da
\]

The distribution $p(x^{\prime} | x, a)$ describes the dynamical model of the environment with $x^{\prime}$ being the next state, $x$ the current state and $a$ the action performed. $p(x^{\prime} | x)$ is the same dynamical model but with the action marginalized out:

\[
      p(x^{\prime} | x) = \int p(x^{\prime} | x, a) p(a|x) da
\]

The channel capacity essentially computes the number of different next states for all possible actions. The channel capacity would be zero if the agent has no control over the environment where every action is leading to the same next state.

Currently the only algorithm used for computing the empowerment value for a single state is the Blahut-Arimoto algorithm \cite{jung_empowerment_2012}.
Both the computation of this KL-divergence and the marginalisation of the system dynamics are very expensive and are done by sampling. Not only does one need to compute these values but also optimize them with respect to $p(a|x)$.
The KL-divergence inside the channel capacity is estimated by monte-carlo integration and then maximised by iteratively changing the probabilities of each discrete action. This is computationally very expensive and not suitable for online use e.g. in a robotic system. In the following we will propose an efficient implementation replacing the Blahut-Arimoto algorithm and enabling the use in real world robotics systems.

\section{Efficient Empowerment}

\subsection{Analytic KL-divergence}

Where in \cite{jung_empowerment_2012} the authors used discrete action distributions and Monte Carlo sampling for computing the empowerment objective we decided to follow \cite{kingma_stochastic_2013} for an efficient computation of the KL-divergence by using the analytical solution for the KL-divergence between two Gaussian distributions \footnote{Obtained with the help of the Q\&A community ``crossvalidated'' at \url{http://stats.stackexchange.com/questions/7440/kl-divergence-between-two-univariate-gaussians}.}. We assume that the system dynamics can be modelled by Gaussian distributions whose parameters are defined by neural networks.

\[
  p(x^{\prime}|x,a) = \mathcal{N}(x^{\prime} | \mu(x,a), \sigma(x,a))
\]

where $\mu(x,a)$ and $\sigma(x,a)$ are modelled by Neural Networks. 

\subsection{Variance Propagation for Marginalisation}

For calculating the Empowerment objective one does not only need the dynamics model $p(x^{\prime}|x,a)$ but also the transition probability $p(x^{\prime}|x)$ with the action $a$ being marginalised out.
Since this marginalisation is very costly we are using a technique called Variance Propagation \cite{wang_fast_2013, bayer_training_2013, bayer_fast_2015}.
Variance Propagation defines a set of rules for transforming a Gaussian when propagating it through a network. By setting the input mean and variance of the action $a$ to the mean and variance of $p(a|x)$ we are effectively marginalising out $a$.

\subsection{Variational Auto-Encoders}

Elements of state and action need to be statistical independent for properly applying Variance Propagation and computing the analytical KL-divergence. Since this does not hold for most real world data we need to transform state and action into latent spaces where their elements are statistical independent.
We are using the Variational Auto-Encoder \cite{kingma_stochastic_2013} to transform state and action into these latent spaces.

\subsection{Action selection}

Empowerment only computes a scalar measuring the quality of the current state. It does not provide suitable actions for controlling a system. The simplest way for creating actions would be to predict the next state given an action using the already available system dynamics and choosing the action producing a next state with highest empowerment.
Another more sophisticated solution would be to use empowerment as the reward function for reinforcement learning. Using it as a regularizer for an already existing reward function is also possible.

\section{Pendulum Experiments}

\begin{figure}
\includegraphics[width=\textwidth]{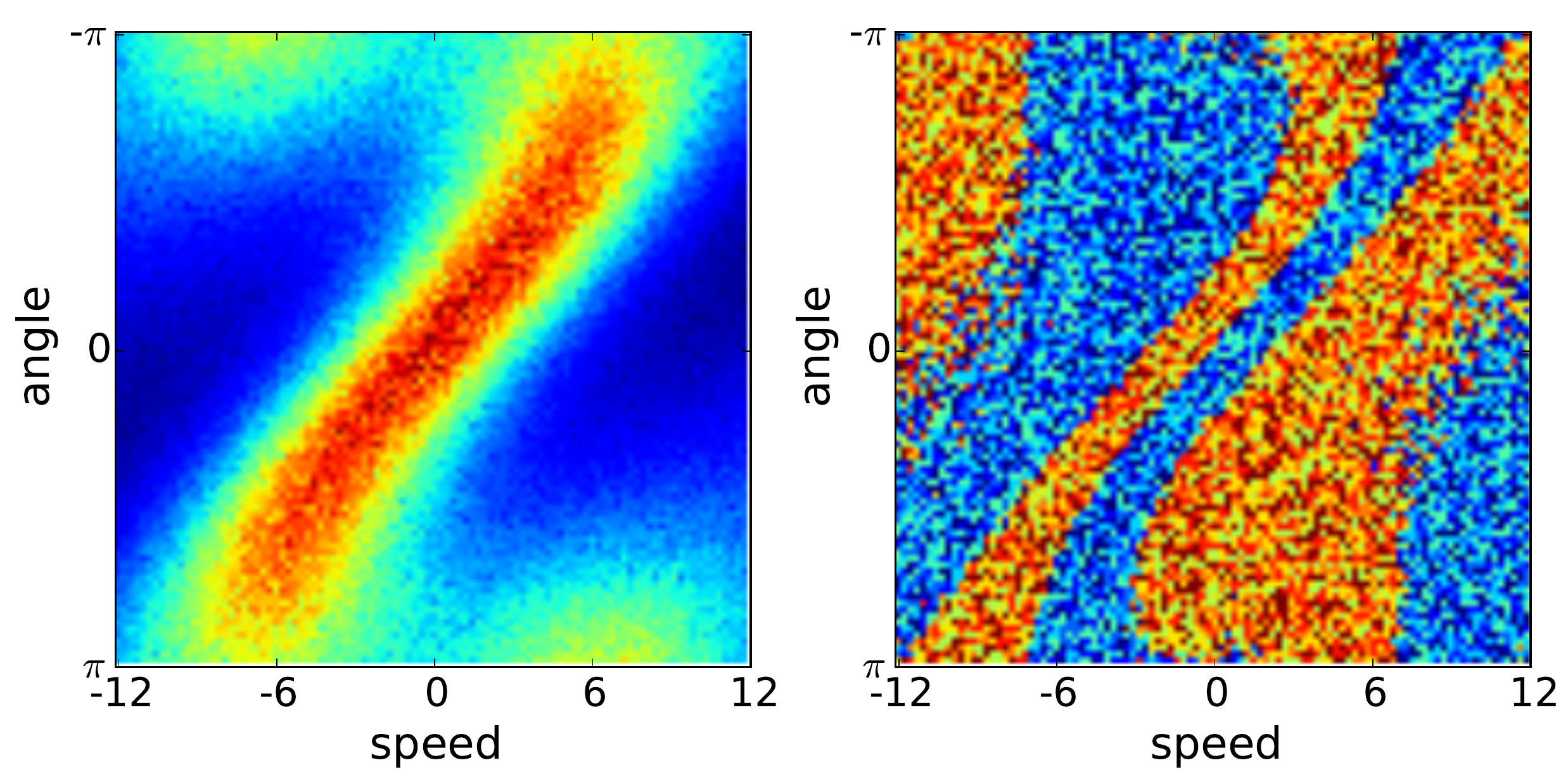}
\caption{Empowerment computed with our efficient implementation on simulated pendulum data. (left) empowerment landscape for different angles and velocities. Red indicates high values and blue low values. (right) chosen action for moving towards states with higher empowerment using the one-step prediction of the system dynamics.}
\label{fig:pend}
\end{figure}

As a first simple experiment we tried our efficient implementation on the pendulum task similar \cite{jung_empowerment_2012}. The system dynamics of this pendulum are known and implemented in a neural network like structure such that we can apply Variance Propagation for integrating out the action. The probability distribution $p(a|x)$ was implemented using a neural network modelling sufficient statistics for a diagonal Gaussian distribution.
In this simple pendulum experiment we did not use the Variational Auto-Encoder trick for making state and action statistical independent. It was not necessary since both elements of the state vector were already independent and the action is only a scalar.
The result of this experiment can be seen in Fig.:~\ref{fig:pend}. The value of empowerment is maximal for the angle and velocity being zero corresponding to the state of the inverted pendulum standing upright.

\section{Conclusion}

We provided a solution for efficiently computing empowerment for high dimensional continuous state and action spaces by combining methods including Variance Propagation, analytical computation of the KL-divergence and the Variational Auto-Encoder. We showed in a first experiment with a simulated inverted pendulum that this method is able to identify states with high empowerment and also able to generate actions using a one-step predictor.

Future work consists of replacing the dynamical model with a learned model. We will also test our algorithm on real world data with high dimensional state and action spaces. Furthermore we plan to test action selection by using  reinforcement learning with empowerment as reward function.

\small{
\bibliographystyle{unsrt}
\bibliography{paper.bib}
}

\end{document}